\documentclass[letterpaper, 10 pt, conference]{ieeeconf}  

\IEEEoverridecommandlockouts                              

\usepackage{graphics} 
\usepackage{epsfig} 
\usepackage{mathptmx} 
\usepackage{times} 
\usepackage{amsmath} 
\usepackage{amssymb}  

\usepackage{comment}
\usepackage{subfigure}

\usepackage[T1]{fontenc}
\usepackage[utf8]{inputenc}  
\usepackage{multirow}

\usepackage{array}
\newcommand{\PreserveBackslash}[1]{\let\temp=\\#1\let\\=\temp}
\newcolumntype{C}[1]{>{\PreserveBackslash\centering}p{#1}}
\newcolumntype{R}[1]{>{\PreserveBackslash\raggedleft}p{#1}}
\newcolumntype{L}[1]{>{\PreserveBackslash\raggedright}p{#1}}

\title{\LARGE \bf    
    Forming a sparse representation for visual place recognition using a neurorobotic approach
}

\author{
	Sylvain Colomer$^{1,2}$,
	Nicolas Cuperlier$^{1}$,
	Guillaume Bresson$^{2}$,
	Olivier Romain$^{1}$
	\thanks{$^{1}$Laboratoire ETIS UMR8051, Université Paris Seine, ENSEA, CNRS, France.}
	\thanks{$^{2}$Institut VEDECOM, 23 bis Allée des Marronniers, 78000 Versailles, France.}
}

\begin{document}
	
\maketitle
\thispagestyle{empty}
\pagestyle{empty}
	
\begin{abstract}

This paper introduces a novel unsupervised neural network model for visual information encoding which aims to address the problem of large-scale visual localization.
Inspired by the structure of the visual cortex, the model (namely HSD) alternates layers of topologic sparse coding and pooling to build a more compact code of visual information.
Intended for visual place recognition (VPR) systems that use local descriptors, the impact of its integration in a bio-inpired model for self-localization (LPMP) is evaluated.
Our experimental results on the KITTI dataset show that HSD improves the runtime speed of LPMP by a factor of at least 2 and its localization accuracy by $10\%$. A comparison with CoHog, a state-of-the-art VPR approach, showed that our method achieves slightly better results.
	
\end{abstract}


\section{Introduction} 

Self-driving car architectures are commonly based on modular systems, where each module is responsible for a sub-task mandatory to achieve autonomous navigation \cite{yurtseverSurveyAutonomousDriving2019}.
Among these different systems, a particular attention must be paid to the self-localization module, potentially one of the blocking points in large-scale deployment of automated vehicles. 

\begin{figure}[h]
	\centering
	\includegraphics[width=0.95\linewidth]{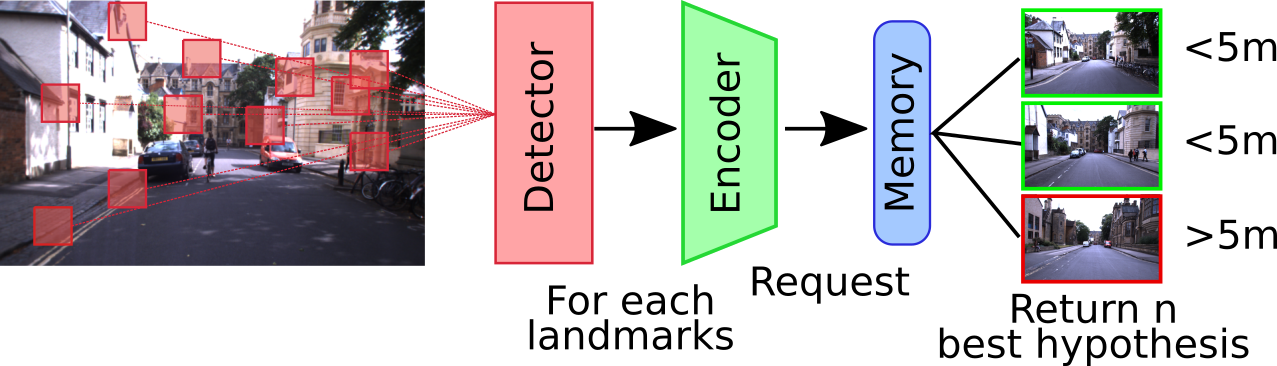} 
	\caption{
    \textbf{General architecture of VPR systems.} VPR systems can be generalized as three functional blocks: a detector, which extracts the important parts of an image, an encoder which compresses and selects the useful information and a memory which contains all the places learned. During a request, the VPR system returns the global coordinates of the place it recognizes best. A request is considered correct if it provides a hypothesis sufficiently close to the position of the vehicle. Depending on the architectures, the distinction between these three functional blocks can be less visible, especially with models which encode the image entirely. 
    }
	\label{figure:SystemeVPR}
\end{figure}

Among the different approaches of self-localization available in the literature \cite{Bresson2017}, the use of vision is gaining more and more interest since cameras are passive and inexpensive sensors that provide a rich flow of information \cite{jessicavanbrummelenAutonomousVehiclePerception2018}. 
This gives rise to \emph{Visual Place Recognition} (VPR) methods aiming at recognizing a location only from the visual appearance of the perceived scene.
Available in many forms such as deep neural networks, bag of words or neuro-cybernetic solutions (see \cite{lowryVisualPlaceRecognition2016} for a survey), these systems often follow a common scheme, starting with an image acquisition followed by some image processing in order to build a representation that best characterizes the current location (figure~\ref{figure:SystemeVPR}).
During the localization phase, this representation is then compared with the current knowledge of the world to output the most probable location. 
A central question thus becomes: what is the most appropriate representation of space to perform the matching process required for localization? This representation has to be both informative enough to discriminate close-looking places with a given accuracy while, at the same time, be compact enough to be processed in real time.

In recent years, great advances in terms of localization performance and computational cost has been made in the field of VPR. 
The arrival of learning models such as convolutional neural networks (CNN) or adversarial models in VPR allowed for higher recognition scores but at the cost of a more or less expensive learning
as reported in \cite{Zaffar_review_2020}.

As seen in the field of artificial vision, CNN models allow to obtain very high performance to process images, in particular to detect and to classify objects \cite{Voulodimos_Doulamis_Doulamis_Protopapadakis}. 
Works like \cite{Sermanet_Eigen_Zhang_Mathieu_Fergus_LeCun_2014} were pioneers in the use of pre-trained CNNs in order to characterize a place and showed that this approach could benefit to the VPR field by providing usable visual features.
Now several models, such as NetVlad, HybridNet or AMOSNet, offer complete solutions to VPR issues and currently give rhe best results in terms of performance on difficult localization conditions \cite{Zaffar_CoHog_2020}. 
However, the downsides of such methods are the computational cost (whether in learning or in use), their need for large learning datasets and their lack of explainability which are important criteria when considering autonomous driving. 
Despite the enthusiasm for deep networks, approaches based on handcrafted features still continue to be proposed, among which one can note CoHog, which exhibits performance scores comparable to CNN models at a much lower cost and without requiring prior training but is still insufficient for accurate large-scale positioning.

Relying on the fact that animals are regularly required to cover large distances in their natural environment \cite{zenoReviewNeurobiologicallyBased2016}, some approaches have developed bio-inspired localization models that imitate key parts of the brain as the hippocampus or the visual cortex \cite{Cuperlier2007,zenoReviewNeurobiologicallyBased2016, Espada2019}. 
Since animals are the only known solution to this positioning problem, studying the mechanisms by which the brain of these biological systems solve it could provide a blueprint to achieve similar efficiency in artificial ones.  

Among these models, sparse coding algorithms (SC) \cite{Spanne_Jorntell_2015} are inspired by the biological system and give a method to compress efficiently information that meet both the previously described requirements (informative and compact). 
By construction, they allow for both a control on the accuracy of the resulting representation and a large encoding capacity by taking advantage of the combinatorics of the few active elements \cite{Olshausen_Field_2004}. 
Its use on localization models is especially interesting as the encoding of visual landmarks was identified as the critical process circumscribing the size of the tackled environment, since the number of learned visual signatures increases with the traveled distance \cite{Espada2019}. 

Our contributions in this paper are the following:
\begin{itemize}
\item We present HSD (Hierarchical Sparse Dictionaries), a bio-inspired neural network allowing to exploit sparse coding of the visual information for automated vehicle localization.
Inspired from the visual processing of mammals, this model proposes to cascade sparse dictionaries to build a lighter representation of a place.
\item We integrated HSD in our bio-inspired localization model called Log-Polar Max-Pi (LPMP) \cite{Espada2019} which relies on the encoding of landmarks in a neural memory.
While being thought as a major improvement of the LPMP model, HSD could also benefit to other VPR systems by providing a compact and robust visual signature of points of interest.
\item We evaluated our method on an open dataset (KITTI) and compared the results with our original approach (LPMP) and with a state-of-the-art VPR method named CoHog \cite{Zaffar_CoHog_2020}. 
\end{itemize}

The rest of this paper is divided as follows: first, we introduce the HSD model and its integration inside our bio-inspired VPR system LPMP, then we present the experimental setup designed to assess the performance of our approach and comment the obtained results. Finally, a short discussion on the contributions is proposed.

\section{Hierarchical Sparse Dictionaries model}

 \subsection{Global overview}

\begin{figure}[h]
	\centering
	\includegraphics[width=1\columnwidth]{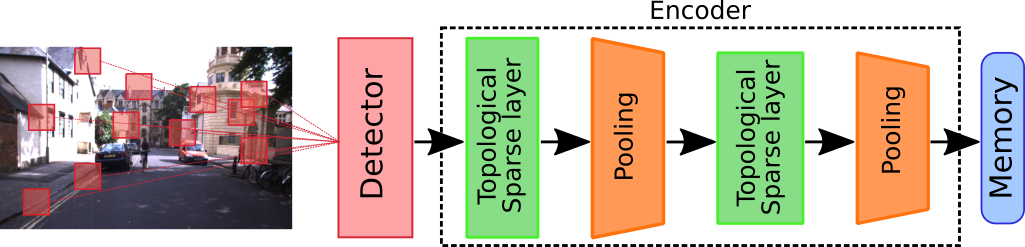} 
	\caption{\textbf{Simplified diagram of the HSD model.} The system encodes landmarks by alternating layers of topologic sparse coding and max pooling, reducing the size of the landmark signatures according to the number of atoms in each dictionary. }
	\label{figure:HSD}
\end{figure}

We introduce in this paper a new unsupervised learning model named Hierarchical Sparse Dictionaries (HSD) which aims to compress visual information in the context of localization for automated vehicles (see figure~\ref{figure:HSD}).
It was designed to encode visual landmarks extracted from the visual scene in a more compact format and to ease their recognition during navigation.  
This compression is carried out by using sparse coding, pooling and topology techniques, respectively known to compress, generalize and structure information.  

\begin{figure}[h]
	\centering
	\includegraphics[width=0.85\linewidth]{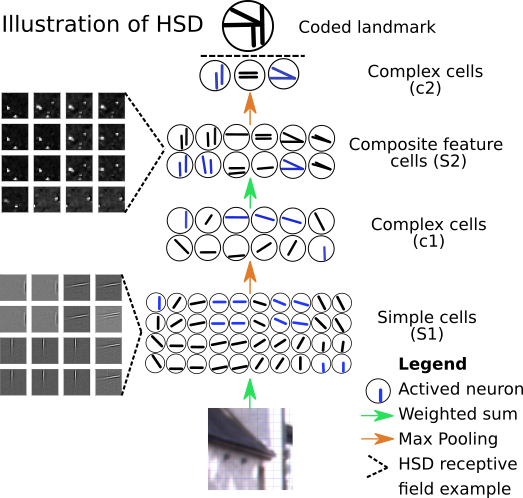} 
	\caption{
    \textbf{Diagram of HSD operation.}
    To encode the image corresponding to a visual landmark, the HSD network learns to decompose it via a cascade of filters. 
    Learning in the S1 layer ends with receptive fields similar to oriented Gabor filters (bottom images on the left).
    The S2 layer neurons learn frequent combinations of co-activated S1 neurons, leading to more complex patterns (top images on the left).
    The neurons of the S1 and S2 layers are arranged spatially in a map thanks to the use of a self-organized map. The topology of the resulting map is such that close neurons code for close features in the input space.
    The intermediate pooling layers C1 and C2 reduce the size of the code by selecting neurons side by side on S1 and S2, which code similar patterns.
    }
	\label{figure:HSD-principle}
\end{figure}

From an architectural point of view, HSD is composed of an alternation of topological sparse layers (TSL) and max-pooling layers. The purpose of TSLs is to efficiently compress incoming information by representing it in a more efficient way. 
However, unlike most sparse codes, we spatially sort the atoms of our sparse layers with a self-organisation algorithm, meaning that they are arranged such that nearby atoms code for closed features in the input space (see figure~\ref{figure:HSD-principle}, layers S1 and S2).
Thus the succession of TSL with max-pooling layers bring together atoms encoding similar features, creating a new generalization property of features which makes the code more robust to translations.

Following a bio-inspired approach, our model coarsely mimics the structure and functional processes of the visual cortex.
The structure of this model intents to replicate the organization of the different visual cortex layers and in particular two kinds of neurons: the simple cells \cite{Olshausen_Field_2004}, sensible to specific orientation, and complex cells \cite{Alonso_Usrey_Reid_2001} that create spatial invariance by applying max pooling mechanisms. 
Simple cells can be modeled by sparse coding dictionaries which build neurons sensible to similar orientation \cite{Olshausen_Field_2004}. Complex cells might be modeled by max pooling layers as in the Hmax model \cite{Serre_Wolf_Bileschi_Riesenhuber_Poggio_2007}, a well-known computational model which aims to reproduce the object recognition capabilities of the mammals visual cortex.
HSD thus proposes a model (see figure~ \ref{figure:HSD-principle} for a diagram of the model) based on simple and complex cells that outlines the processes performed by the visual cortex when the mammals brain is engaged in a place recognition task. 

\subsection{Basic sparse layer}

Being a direct reading of the efficient coding theory \cite{Spanne_Jorntell_2015}, which conjectures that the brain synaptic communication optimizes criteria of metabolic efficiency, sparse coding algorithms allow to build an efficient representation of input data by optimizing the following criterion: 
 \begin{equation}
  \boldsymbol{s}=\sum_{m=1}^{M} a_{m} \boldsymbol{\phi}_{m} \quad \
    \text{subject to} 
  \quad \min _{\boldsymbol{a}}\|\boldsymbol{a}\|_{0} 
  \label{eq-sc}
 \end{equation}where $s\in \mathbb{R}^{n}$ represents the input vector, $a_{m}$ an activity coefficient, $\phi_m \in \mathbb{R}^{N}$  an element of the dictionary $\Phi$ and $\|.\|_{0}$ the $L_0$ norm. 
In HSD, sparse layers are learned using Homeostatic Sparse Hebbian Learning \cite{Perrinet_2010}, an iterative algorithm designed to build a sparse dictionary on natural images. 
This algorithm builds a sparse dictionary by alternating two processes: an update step where the dictionary is modified to improve reconstruction, and an encoding step where the current input is approximated with a limited number of atoms.
The goal of the two-step process is to optimize the image cost, defined using a Generative Linear Model\footnote{A LGM model is a model where we consider that an image $I$ can be expressed as: $I=\Phi^{T}\mathbf{a}_{k}+\epsilon$ with $\epsilon \in \mathbb{R}^{M}$ as Gaussian noise} as: 
\begin{equation}
\mathcal{C}(\mathbf{a} \mid \mathbf{I}, \Phi)=\frac{1}{2 \sigma_{n}^{2}}\|\mathbf{I}-\Phi \mathbf{a}\|^{2}+\lambda\|\mathbf{a}\|_{0}
\label{eq-sc-cost}
\end{equation}where $\sigma$ is the signal scaling steepness, $\lambda$ is the number of bits which weights the norm $L_0$ and $I$ is the input image (more details are given in original publication \cite{Perrinet_2010}). 
This cost equation expresses the same idea as Equation (\ref{eq-sc}) where the first term corresponds to the reconstruction cost and the second term is the representation cost or sparse cost.
The encoded images are filtered beforehand by applying whitening techniques as in \cite{Olshausen_Field_2004}.

\subsubsection{Update of dictionary} The dictionary update is carried out with an an Hebb's rule \cite{Kuriscak_Marsalek_Stroffek_Toth_2015}, formulated for an atom $a_i$ as: 
\begin{equation}
\phi_{i} \leftarrow \phi_{i}+\eta a_{i}(\mathbf{I}-\phi \mathbf{a})
\label{eq-sc-update}
\end{equation}where $\eta$ is the learning rate of the gradient descent. Thus this formula makes it possible to strengthen the atoms which were used to reconstruct the signal, knowing that the system can only use a limited number of atoms to reconstruct the signal. 

\subsubsection{Encoding of signal} To perform the encoding of a signal, HSD uses an homeostatic version of the Matching Pursuit algorithm \cite{Mallat_Zhifeng_Zhang_1993}.
The goal of this algorithm is to recursively minimize the global reconstruction error between the input signal and the reconstructed signal by pondering the atoms that improve reconstruction. 
At a given iteration $t$, the system tries to improve its intermediate representation (expressed by $ \hat{a}_{i}$) and seeks to reduce the intermediate reconstruction error named the residue $r(t)$:
\begin{equation}
r(t)=I-\sum_{i} \hat{a}_{i}(t) \phi_{i}
\label{eq-sc-residu}
\end{equation}It thus seeks out the best combination of atoms and activities to minimize the residue knowing the atoms already activated, $i^{*}$ representing the index of the selected atom: 
\begin{equation}
i^{*}=\operatorname{ArgMax}_{i}\left[z_{i}\left(\hat{a}_{i}\right)\right]
\label{eq-bestAtom}
\end{equation}
To make the atoms of the dictionary homogeneous, the value of $\hat{a}_{i}(t)$ is transformed by $z(.)$, a homeostasis function described by the equation \ref{eq-sc-homeo}.
The function $z(.)$ modifies the activity of the sparse code by equalizing it so that the final dictionary remains homogeneous.
This treatment, close to a histogram equalization, requires the computation of $dP_i$, the probability distribution of the variable $a_i$.
Since this distribution cannot be known upstream, it must be updated at each learning step, as described below: 
\begin{equation}
z_{i}\left(\hat{a}_{i}\right) \leftarrow\left(1-\eta_{h}\right) z_{i}\left(\hat{a}_{i}\right)+\eta_{h} P\left(a_{i} \leq \hat{a}_{i}\right)
\label{eq-sc-homeo}
\end{equation}
The encoding step is stopped when the maximum number of authorized atoms $N_{0}$ is reached, a parameter which indirectly controls the level of sparsity of the final code. 

\subsection{Topological sparse coding layer}

\begin{figure*}[h!]
	\centering
	\includegraphics[scale=0.8]{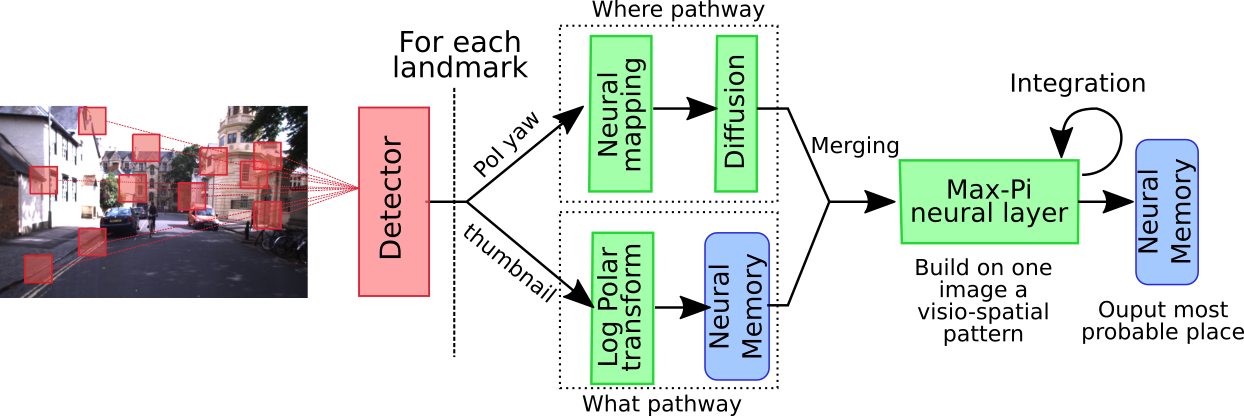}
	\caption{
    \textbf{Functional diagram of the LPMP model.}
    To learn a place, the system builds a visio-spatial pattern by accumulating on the Max-Pi neural layer, the visual identity of each landmark with its absolute angle. 
    These two information built in parallel are extracted for each landmark via two blocks: a log-polar block which transforms, encodes and stores in a memory each landmark; a spatial block which encodes the absolute angle of the landmark in a neuron vector. 
    }
	\label{figure:LPMP}
\end{figure*}

The application of HSHL on landmarks extracted from an image leads to the construction of an orientation sensitive dictionary which resembles a rich Gabor dictionary. However, the atoms in this dictionary are not sorted, which means that atoms encoding a similar pattern are not placed close to each other. Thus the code generated by this dictionary is very sensitive to the position of the encoded landmark. Indeed, a small translation of the landmark in the image can induce a strong change in the activity pattern of the sparse code layer, making its use in the matching process of a VPR system impossible.  

Pooling operations are commonly used to achieve translational invariance in convolutional networks as in the Hmax algorithm \cite{Serre_Wolf_Bileschi_Riesenhuber_Poggio_2007}, where this problem has been solved by applying a max-pooling mechanism between Gabor filters of the same orientation.
But a similar pooling method cannot be directly applied in HSD since its sparse dictionaries result from an unsupervised process where no constraint is given on the spatial arrangement of neurons having to code for a close entry. It is consequently impossible to know which neuron to pool.

In the visual cortex, several experiments have shown that simple cells are "sorted" layer by layer. For example, the V1 cortex has a topology where neurons sensitive to similar orientations are side by side \cite{Paik_Ringach_2012}. 
This property is one of the keys to the visual cortex which allows for the building of increasingly complex filters by merging information from lower levels thanks to complex cells.

In HSD, a Kohonen network is used to sort our dictionaries \cite{kohonen2012self} leading to a new kind of layer: a topological sparse layer (TSL).
The Kohonen network or Self-Organizing Map (SOM) is a well-known neural network generally used for dimension reduction of the input space. It leads to the formation of a discrete map, usually in 2D, that preserves the topological properties of the input space by using competitive learning and a neighborhood function.
To update the network at an instant $t$, two steps are carried out:
the first stage is the competition where each neuron competes to represent the entry $x_i$. The winner neuron $c(t)$ is designed as the best matching neuron:
\begin{equation}
c(t)=\underset{i}{\arg \min }\left\{\left\|x(t)-\phi_{i}(t)\right\|\right\}
  \end{equation}
The second step consists in updating the weights $w_{i}$ of the winning neuron and its closest neighbors using: 
\begin{equation}
w_{i}(t+1)=w_{i}(t)+\alpha(t) h_{c i}(t)\left[x(t)-w_{i}(t)\right]
\end{equation}The membership of a neuron in the neighborhood of the winner is defined by the function $h_{c i}(t) $, a Gaussian function which decreases with the value of the distance in the map to the winning neuron. The learning rate is controlled by $\alpha(t)$.

Reconstructing a topology from the sparse dictionaries makes it possible to apply a max-pooling operation to achieve translational invariance which helps in recognizing previously seen landmarks in a moving vehicle.

Cascading topological sparse dictionaries makes possible the compression of visual information into an efficient representation. This hierarchical organization even improves the localization performance of the system by generalizing redundant structures in visual scenes.

\section{Integration of HSD in a neuromimetic approach}

While there are many methods for performing visual localization, few of them are inspired by the functioning of mammals \cite{zenoReviewNeurobiologicallyBased2016}.
Following a neuromimetic approach, the LPMP model (Log-Polar Max Pi) \cite{Espada2019} aims at offering a new pipeline to solve VPR issues.
LPMP relies on a neural network model (see figure~\ref{figure:LPMP}) which combines several key theories on how biological systems may build a neural map of the environment. 
More precisely, LPMP aims to model the spatial neurons observed in the animal hippocampus called \emph{place cells}.
Like biological ones, the simulated place cells respond with a high firing rate for a given place in the environment, named place field, and has shown interesting property of robustness in complex environments \cite{Cuperlier2007, Espada2019}. 

In this model, two information flows for each detected landmark are processed in parallel: 
\begin{enumerate}
    \item The \textit{visual identity} (or \textit{what pathway}), which corresponds to the encoding of a local view centered on extracted points of interest (PoIs) in the image provided by a detector\footnote{PoIs are extracted on a saliency map built with a Deriche filter and a Difference of Gaussians filter.}. 
    \item The \textit{azimuth information} (or \textit{where pathway}) computed from the vehicle absolute orientation \footnote{Information which might be obtained from a magnetic or visual compass \cite{vis_compass} or directly derived from the GPS of the vehicle} combined with the PoI coordinate in the image. The \textit{azimuth information} is convoluted with a Gaussian kernel. The spreading of this information allows its generalization to close values.
\end{enumerate}
For each PoI, both types of information are then merged and accumulated in a neural layer (named \textit{Max-Pi neural layer}), whose neurons activity is characteristic of the current position of the vehicle (see \cite{Espada2019} for a more detailed description of this process).
Through unsupervised learning, visual place cells (VPC) then simply result from the learning of the activity pattern of this \textit{Max-Pi neural layer} while the vehicle explores an unknown environment.

Contrary to most classic VPR systems which rely on a fixed step to learn new places, this task is performed in the LPMP model thanks to a novelty detector called the \textit{vigilance loop}. This mechanism triggers the learning when the most activated VPC falls below a specific recognition threshold. Consequently, this parameter can be used to control the mean size of the generated place fields.

We integrate our proposed sparse coding model HSD inside LPMP on order to evaluate its impact in terms of localization performance. To do so, the log-polar transform responsible for the landmark encoding of the LPMP model (see figure~\ref{figure:LPMP}) is replaced by HSD. The rest of the LPMP model remains unchanged. 
This new version of the model is called in the rest of the paper HSD+MP.

\section{Experiments}

\subsection{Datasets} \label{sec:datasets}

We validate our approach HSD+MP in the context of automated driving using the KITTI dataset \cite{Geiger_Lenz_Stiller_Urtasun_2013}, a commonly used dataset in VPR.
In order to assess the ability of the tested VPR models to recognize previously visited locations, a series of three routes are extracted from the dataset (details of the cut sequences are shown in Table \ref{tab:kitti-split}). These routes were specifically selected because the vehicle revisits each location twice, thus offering two distinct trajectories. One was used for the \textit{learning phase} (sequence 1 in Table \ref{tab:kitti-split}), the other for the evaluation of the models named the \textit{test phase} (sequence 2 in Table \ref{tab:kitti-split}). 
Due to the small size of the KITTI dataset (in terms of places revisited), these routes only belong to two types of environment: wide roads with a large and deep field of view and narrow streets.
The images of each sequence were reduced to 642x188 pixels.

\begin{table}[!ht]
	\centering
	\caption{Summary of the used KITTI sequences.} 
	\begin{tabular}{|c|c|c|c|}
	\hline
	\textbf{Route id} & \textbf{K0-1} & \textbf{K5-1} & \textbf{K5-2}\\
    \hline
    Dataset name &KITTI 00 & KITTI 05 & KITTI 05\\
    \hline 
    Indices of sequence 1 &392-932 &10-116 & 550-780\\
    \hline 
    Indices of sequence 2 &3399-3839 &2420-2511 & 1289-1554 \\
    \hline 
    Distance & 378m & 96m & 199m\\
    \hline 
    Environment & Mixed & Large roads & Narrow streets\\
    \hline
	\end{tabular}
	\label{tab:kitti-split}
\end{table}

For the learning of the sparse dictionary, we built a separate dataset of landmarks via the detector of the LPMP model. 
This dataset was extracted from the first $150$ images of $KITTI 00$, $KITTI 05$, apart from the sequences with revisited places used for the \textit{learning phase} and the \textit{test phase}. 

\subsection{Evaluation methodology}\label{sec:Evaluation}

To confirm the contribution of HSD, two experiments on LPMP and HSD+MP were carried out: first, the influence of the HSD configuration (HSD network setting) on computational and localization performances of HSD+MP was measured on $K5-1$.
Second, the performances of the HSD+MP model on all the routes were assessed with the best configuration of HSD (as determined in the previous test). 
In order to situate its performance with respect to the state of the art, the HSD+MP model is evaluated on the same experimental conditions against the VPR model named CoHog \cite{Zaffar_CoHog_2020}, one of the lightest network available with one of the best localization scores \cite{Zaffar_review_2020}. 

\subsubsection{Localization performance} 
The ability of each model to recognize places already visited by the vehicle was measured. To do so, standard precision/recall measures, summarized with their area under the curve (AUC) were used. The places to be learned were selected on the basis of the distance traveled, calculated using the ground truth (DGPS coordinates). For each tested trajectory, three experiments were carried out with images sampled according to three different distances: 2m, 3m and 5m.  
For the LPMP and HSD+MP models, whose learning is controlled by a vigilance loop, different recognition thresholds were explored to obtain, on average, a similar sampling rate of places.
To compensate GPS synchronization errors between the two trajectories of the same route, a tolerance, specific for each location, was added, on the basis of the GPS imprecision (on average $0.79m$ with a standard deviation of $0.98m$ ).

\subsubsection{Computational cost} 
The evaluation of computational cost is carried out by measuring the average processing time of an image for each model after learning. This measurement is converted in frequency to give an idea of the achievable computational frequency in real time.
Different tests were carried out on a single CPU (no GPU) to limit the impact of the code optimization level of each model. All the experiments were carried out using an AMD Ryzen Threadripper 2990wx (3.7GHz) and 64Go of RAM. 

\subsubsection{VPC encoding field} 
\begin{figure}[htbp]
	\centering
	\includegraphics[scale=0.47]{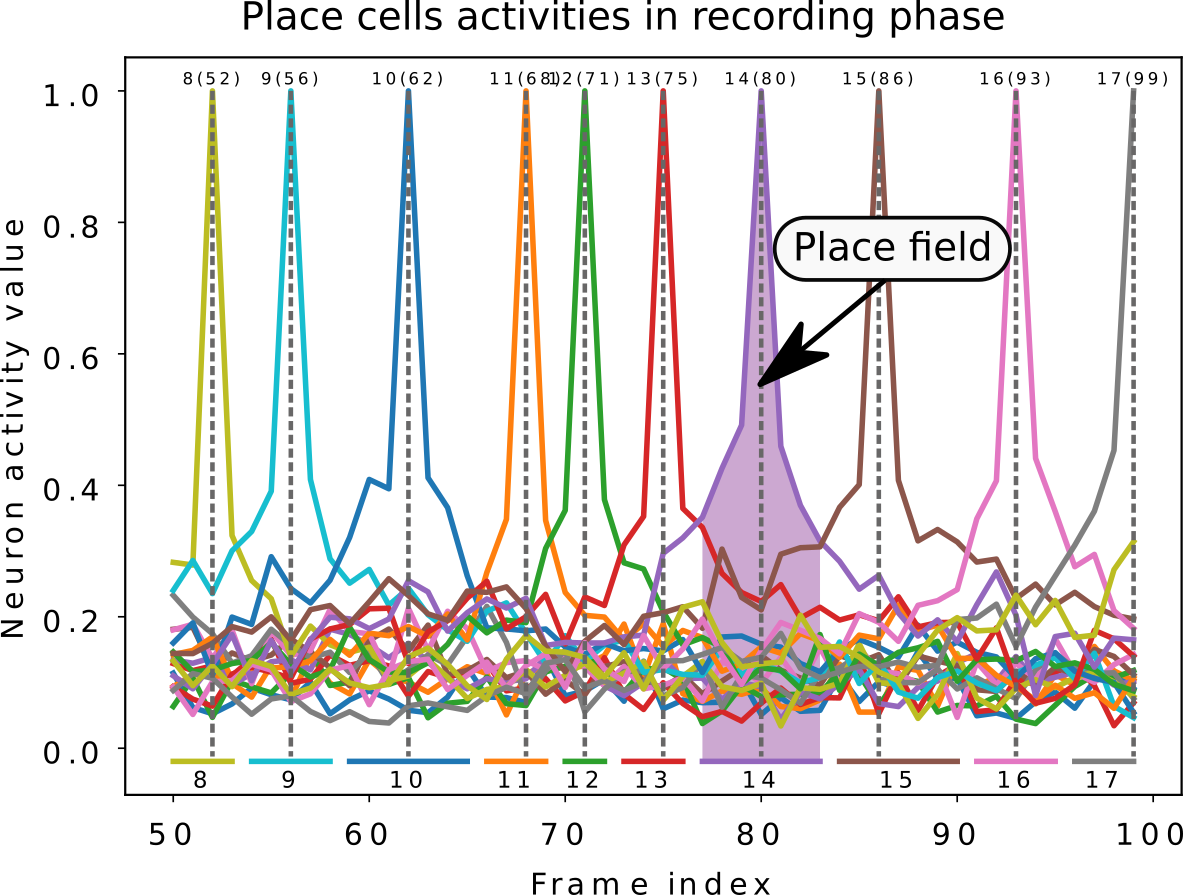} 
	\caption{\textbf{Illustration of a recording step.} 
    During this step, the responses of all VPCs of LPMP or HSD+MP are recorded for each image of the \textit{learning sequence}. No learning occurred during this phase which is only performed to record which cell should respond to a given position in order to build the ground truth used to measure the localization performance when images of the \textit{test sequence} were processed.
    This way, it was possible to measure which space codes a VPC neuron around its point of creation (indicated by the dashed gray lines), as illustrated in purple. The number on top of the dashed gray lines denotes the VPC index and the number into bracket indicates the index of the corresponding image in the sequence.
     }
	\label{figure:placeField}
\end{figure}

The computation of a precision-recall curve on a localization model is commonly done by applying a threshold distance between the current coordinates and the coordinates of the image recognized by the VPR model.
However, due to the causal nature of learning and since a route is explored in only one direction, only the second half of each place field can be observed (from the learning point up to the last location where the cell fires). A second pass is therefore necessary to be able to measure complete place fields (encompassing the first half corresponding to the response of a cell before reaching the precise learning point). To solve this problem, an intermediate phase is performed between the learning and the test phases. This is called the \textit{recording phase} where all the images of the \textit{learning sequence} are fed into the network for a second time (but without any learning) in order to record the full place fields formed during the learning phase and to measure their sizes, as illustrated on figure ~\ref{figure:placeField}. This record points out which cell should respond on a given position and allowed us to build the ground truth used to measure the localization performance when images of the \textit{test sequence} were processed.

\subsubsection{Parameters} 
The parameters of the LPMP and HSD+MP models are the same used in \cite{Espada2019}.
In the particular case of the HSD+MP model, the dictionary is pre-learned on the landmarks dataset as explained in \ref{sec:datasets}.
The S1 layer was learned with a constraint $N_0 = 10$ and the C2 layer was learned with a constraint $N_0 = 5$. The pooling used between S1 and S2 and between S2 and the last layer was of size $2$ and the reorganisation was made with the help of a Kohonen network of neighborhood set to $4$. 

\subsection{Results} \label{sec:Results}
The experiments were carried out first to quantify the contribution of HSD to the original LPMP model, then to evaluate in what extent it could make the model more competitive over some VPR solutions of the state of the art.
To facilitate the notation of the HSD network configuration (number of neurons used by layer) in the result part,  the following convention has been adopted: for a balanced configuration of HSD (all layers are of the same size), the model was called $HSD-n$ with $n$ the first dimension of the layer. Thus HSD-15 is composed of layers of $15*15$ neurons.

\subsubsection{Evaluation of configuration/performance \label{sec:encoding}}

\begin{figure}[h]
	\centering
	\includegraphics[width=0.9\linewidth]{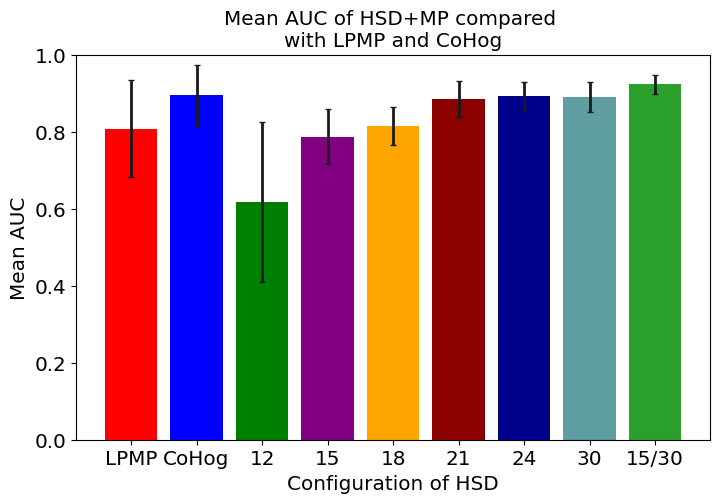}
	\caption{
    \textbf{Comparison of HSD configurations on localization performances.}
    The graph presents the mean AUC of precision-recall curves of HSD+MP, LPMP and CoHog computed with different place sampling rates on K5-1.
    The standard deviation is indicated by a horizontal black bar.
    }
	\label{figure:AUC_HSD}
\end{figure}

\begin{figure}[h]
	\centering
	\includegraphics[width=1\linewidth]{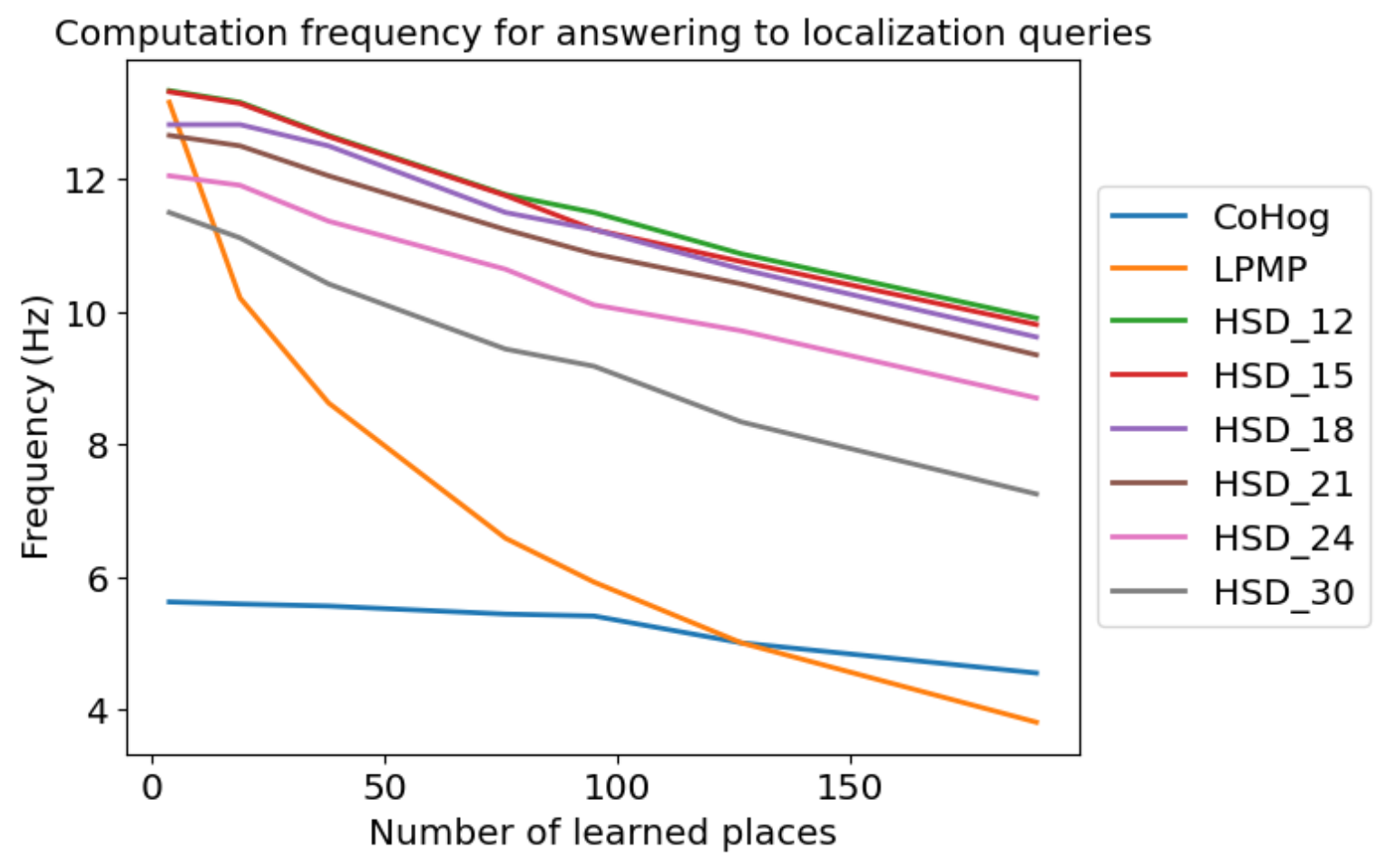}
    \caption{
	 \textbf{Frequency achieved for answering to localization queries.}
     This graph presents the frequency at which LPMP, different configurations of HSD+MP and CoHog answer to localization queries depending on the number of locations learned at a equal place sampling rate on the $K5-1$ route.
    }
	\label{figure:freq_graph}
\end{figure}

Figures \ref{figure:AUC_HSD} and \ref{figure:freq_graph} present the influence of six HSD configurations on HSD+MP and provide a comparison with LPMP. Localization performances of one configuration are estimated by calculating the mean value of precision-recall AUC for the three distances values used to sample places. These tests were performed on the dataset $K5-1$.
Among these configurations, five are balanced and one (called the HSD-15/30) is unbalanced. The $HSD-15/30$ is a specific configuration where the size of the Kohonen network is twice as large as the learned sparse code.

\begin{figure}[h]
	\centering
	\includegraphics[width=0.9\linewidth]{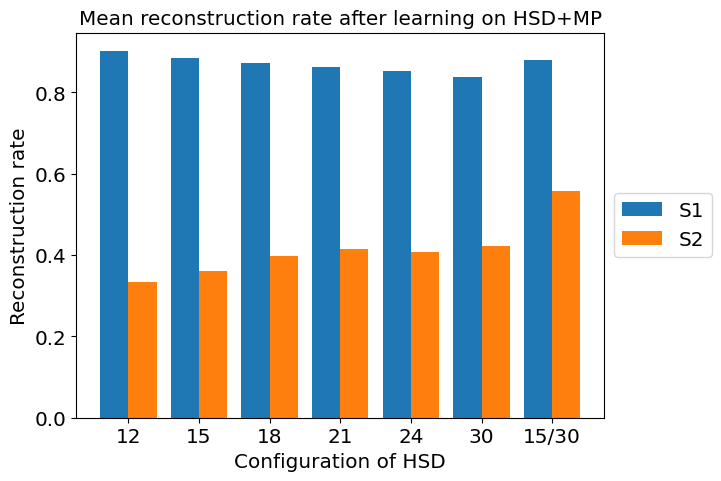}
    \caption{
	 \textbf{Evaluation of the HSD mean reconstruction rate after learning.}
     To learn a representation, layers S1 and S2 seek to optimize the reconstruction of the input landmark.
     This graph shows the average reconstruction rate achieved after learning a dictionary for each HSD configuration.
    }
	\label{figure:graphe_reconstruction}
\end{figure}

Figure \ref{figure:AUC_HSD} shows that increasing the number of atoms in HSD tends to improve the localization performance of the HSD+MP model. 
Going from HSD-18 (81 features) to HSD-24 (144 features) has improved the localization performance by $10\%$.
From HSD-21 to HSD-30, a plateau phase was reached where the place recognition performance remained stable at an average mean AUC value of $85$. 
This plateau is explained by a drop in the maximum reconstruction rate obtained by the S2 layer of HSD, detailed in the figure \ref{figure:graphe_reconstruction}.
Using an unbalanced configuration like HSD-15/30 has enhanced reconstruction, which results in improved localization accuracy. 

However, this improvement has increased the computation frequency of HSD + LPMP.
Figure \ref{figure:freq_graph} shows that the increase in the size of HSD was accompanied by a decrease in the computation frequency.
Thus going from HSD-18 to HSD-24 leads to a decrease in the computation frequency by $9\%$.

Comparing with LPMP at equal performance, HSD-18 allows to achieve a gain of $*2$ on the computation frequency for $150$ learned locations. 
For the best performing configuration HSD-15/30 (which has the same computation time than HSD-30 ), the computation frequency is $80\%$ higher than LPMP for a $10\%$ improvement in performance .
Between $0$ and $15$ learned places, LPMP is less expensive than HSD+MP because the computational cost of the neural network is more important than the computation of the polar log.
However, after 15 places, the gap of the computational memory cost widens resulting in a drastic reduction in computing frequency for the LPMP model.

By equalizing the score with CoHog, \textit{i.e.} using the configuration HSD-24,  HSD+MP has a frequency also twice as high as CoHog. 
For example, HSD+MP can run at a frequency of $10.5 MHz$ for $100$ learned places against $5.5MHz$ for CoHog.
Taking HSD-15/30, the model is $41\%$ faster than CoHog, yet considered to be one of the lightest state-of-the-art neural networks, for an improvement of $4\%$ in the localization performance.

\subsubsection{Global evaluation of performance}
The following experiments were done on the all the routes using only the best configuration (HSD-15/30) retained from the previous tests.

\begin{figure}[ht]
	\centering
	\includegraphics[width=1\linewidth]{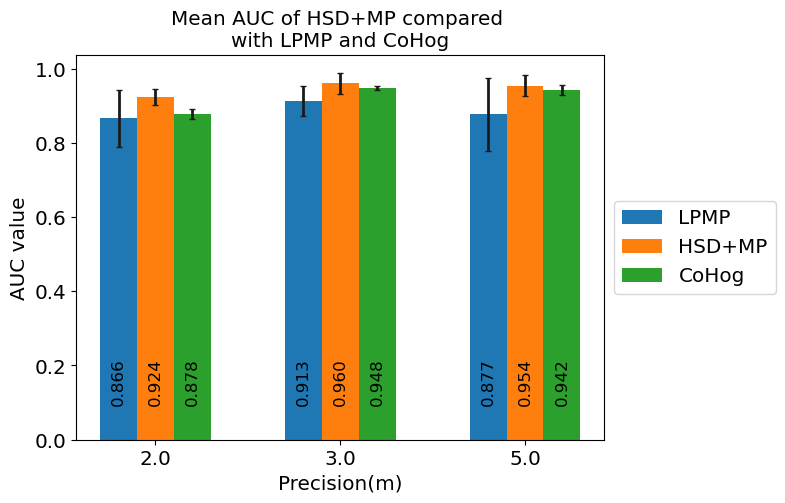}
	\caption{
    \textbf{Evaluation of the mean performance for different place sampling rates}
    The graph presents the mean AUC of precision-recall curves computed on all routes with the three different place sampling rates. 
    The standard deviation is indicated by an horizontal black bar.
    }
	\label{figure:result_global}
\end{figure}

Graph \ref{figure:result_global} shows that HSD has improved the score of LPMP by $7\%$ in average for all the evaluated place sampling rates. More particularly, when places were sampled every 5m, where the generalization of the code was essential, the AUC of HSD+MP was $8\%$ better than LPMP (against $5\%$ at 2m).
Regarding the comparison between CoHog and HSD+MP, we can see that both models give very similar results in accuracy, with a slight advantage for HSD+MP.

\begin{figure}[h!]
	\centering
	\includegraphics[width=1\linewidth]{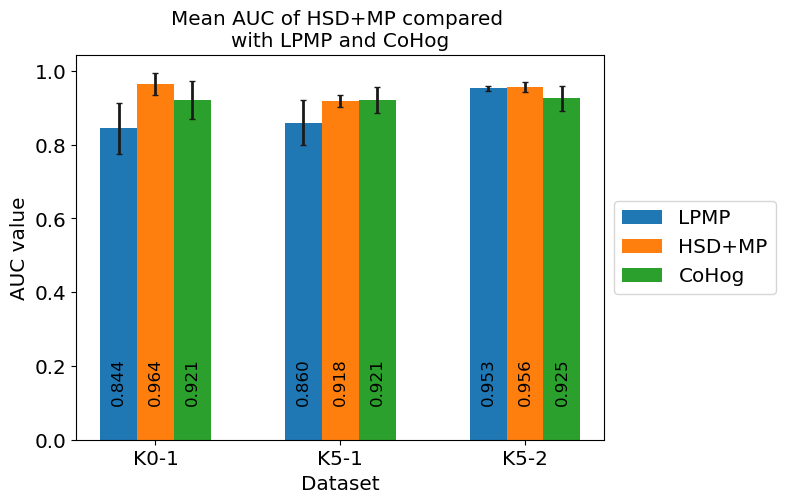}
    \caption{
    \textbf{Evaluation of the mean performance on each sequence}
    The graph presents the mean AUC of precision-recall curves computed for each route on all place sampling rates.
    The standard deviation is indicated by a horizontal black bar.
    }
	\label{figure:result_env}
\end{figure}
Graph \ref{figure:result_env} underlines that HSD+MP has more stable AUC values over the different routes, as shown by the smaller standard deviation (on average twice lower than LPMP and even CoHog).
This graph also reveals strong disparities between the responses of LPMP and CoHog which gave very variable results between $K1-1$ and $K5-1$ routes.

\subsubsection{Learning time and computational cost}
Table \ref{table:cost} focuses on the encoding part of the LPMP and HSD models and provides the estimation of their computational costs (memory consumption) and the learning time of the pre-training phase (for HSD only). Regarding the learning time of HSD, we see that it increases strongly according to the size of the dictionaries learned. 
Thus using a configuration such as the 15/30 strongly reduces the learning time (with similar performances as seen before). 
This unbalanced configuration has the advantage of having the same calculation time as HSD-30, but can be learned twice as quickly. 

\begin{table}[h!]
	\centering
	\caption{Evaluation of the architecture costs}
	\begin{tabular}{
	|c|
	c|
	c|
    c|
    c|
	}
	\hline 
	Model & Learning time &  PoI descriptor\\
     &  &  size \\
	\hline 
    LPMP & . & 2916\\
    HSD-12 & 132s & 36\\
    HSD-15 & 262s & 64\\
    HSD-18 & 313s & 81\\
    HSD-21 & 555s & 100\\
    HSD-24 & 555s & 144\\
    HSD-30 & 1527s & 225\\
    HSD-15/30 & 673s & 225\\
    \hline 
	\end{tabular}
    \caption{
    \textbf{Computational cost and learning time of HSD.}
    This table provides two measurements about the encoding part of LPMP and HSD. The first column shows the learning time of the pre-training phase for HSD and the second column gives the number of elements that make up the PoI descriptors.  
    }
	\label{table:cost}
\end{table}

Regarding the size of the encoded features, the transition to HSD+MP goes from a code of 2916 to 225 in the best configuration. 
This significant reduction in the size of the code of a landmark signature explains the significant gain in computing time and also induces a lower memory usage.

\section{Discussion and conclusion}

In this paper, we underlined the interest of sparse coding and its necessary adaptation to apply it to the navigation problem. We proposed to use a sparse code on visual localization by building a multi-layer neural architecture.
Experimental results showed the advantage, be it on the computation cost or the localization accuracy, of the proposed method (HSD) when integrated into our bio-inspired localization model (LPMP) by replacing its original landmark encoding method. 
The resulting model (HSD+MP), by taking into account more precisely some key properties of the visual processing of mammals, can be seen as a major improvement of LPMP, enabling it to tackle larger environments.
More generally, given that the proposed HSD method provides a PoI signature encoded as a simple vector of floating point values, it could easily be integrated and possibly benefit other VPR systems.   
The results showed a comparable, and even slightly better, localization accuracy than a state-of-the-art model, CoHog. 

The primary goal of our work was not to improve the localization performance of the model but to compress the visual information, determined by previous work as the blocking point of the system. 
However, our results showed that HSD has the potential to be used as a powerful encoding system for visual information. 
It might be possible to keep pushing the performance of HSD by increasing the number of layers, by using pyramidal pooling layer as in CNN or by increasing the $L_0$ norm used in the $S2$ layer. 
However, these solutions lead to a global increase in the computational cost of the system or in the learning time of the network in the pre-training phase, which defeats the objective of using the HSD model as a light compression architecture for autonomous vehicles.  
Future works are therefore planned to study to what extent the results can be improved and assess them on more challenging datasets like the Oxford dataset \cite{RobotCarDatasetIJRR}.

The proposed architecture although sharing similar concepts with Hmax, is rather different in its conception. Indeed, the HSD model does not use convolution but just a classic propagation, which is also much less expensive. 

It is also possible, to some extent, to find common ground between HSD and deep networks.
If both HSD and CNN based models are feed forward neural models and need a pre-training phase, the main difference is about the nature of the training set and the learning time. 
CNN based models require huge labeled datasets, whereas HSD only needs a subsample of a small dataset. 
In addition, the learning time is very short and the learning is unsupervised.

With a significant increase of localization and computation performances when integrating HSD in LPMP, the frequency reached meets the real-time constraints necessary to make the HSD+MP model work online in a vehicle evolving in large environments.
However, there are still many avenues for improvement.
First, this paper focused on the interest of relying on a sparse code for the encoding of visual landmarks which is only a small part of the full VPR model. 
The rest of the HSD+MP model remains identical to the original LPMP model that does not benefit yet of a sparse encoding (and so computational gain).
Second, the current model implementation has not been developed with computation optimization in mind and thus the presented computation time might be quite far from the optimal one.
Work was thus undertaken to study these questions, in particular by looking at a hybrid hardware (SoC) implementation \cite{elouaret:hal-02532875}. 

\addtolength{\textheight}{-8cm}   





\bibliographystyle{IEEEtran}
\bibliography{IEEEabrv, reference}

\begin{thebibliography}{10}
\providecommand{\url}[1]{#1}
\csname url@rmstyle\endcsname
\providecommand{\newblock}{\relax}
\providecommand{\bibinfo}[2]{#2}
\providecommand\BIBentrySTDinterwordspacing{\spaceskip=0pt\relax}
\providecommand\BIBentryALTinterwordstretchfactor{4}
\providecommand\BIBentryALTinterwordspacing{\spaceskip=\fontdimen2\font plus
\BIBentryALTinterwordstretchfactor\fontdimen3\font minus
  \fontdimen4\font\relax}
\providecommand\BIBforeignlanguage[2]{{%
\expandafter\ifx\csname l@#1\endcsname\relax
\typeout{** WARNING: IEEEtran.bst: No hyphenation pattern has been}%
\typeout{** loaded for the language `#1'. Using the pattern for}%
\typeout{** the default language instead.}%
\else
\language=\csname l@#1\endcsname
\fi
#2}}

\bibitem{Alonso_Usrey_Reid_2001}
J.-M. Alonso, W.~M. Usrey, and R.~C. Reid, ``Rules of connectivity between
  geniculate cells and simple cells in cat primary visual cortex,'' \emph{The
  Journal of Neuroscience}, vol.~21, no.~11, p. 4002–4015, Jun 2001.

\bibitem{Bresson2017}
G.~Bresson, Z.~Alsayed, L.~Yu, and S.~Glaser,
  ``\BIBforeignlanguage{en}{Simultaneous {Localization} and {Mapping}: {A}
  {Survey} of {Current} {Trends} in {Autonomous} {Driving}},''
  \emph{\BIBforeignlanguage{en}{IEEE Transactions on Intelligent Vehicles}},
  vol.~2, no.~3, pp. 194--220, Sept. 2017.

\bibitem{Cuperlier2007}
N.~Cuperlier, ``\BIBforeignlanguage{en}{Neurobiologically inspired mobile robot
  navigation and planning},'' \emph{\BIBforeignlanguage{en}{Frontiers in
  Neurorobotics}}, vol.~1, 2007.

\bibitem{elouaret:hal-02532875}
T.~Elouaret, S.~Zuckerman, L.~Kessal, Y.~Espada, N.~Cuperlier, G.~Bresson,
  F.~B. Ouezdou, and O.~Romain, ``{Position Paper: Prototyping Autonomous
  Vehicles Applications with Heterogeneous Multi-FpgaSystems},'' in \emph{{2019
  UK/ China Emerging Technologies (UCET)}}.\hskip 1em plus 0.5em minus
  0.4em\relax Glasgow, United Kingdom: {IEEE}, Aug. 2019, pp. 1--2.

\bibitem{Espada2019}
Y.~Espada, N.~Cuperlier, G.~Bresson, and O.~Romain,
  ``\BIBforeignlanguage{en}{From {Neurorobotic} {Localization} to {Autonomous}
  {Vehicles}},'' \emph{\BIBforeignlanguage{en}{Unmanned Systems}}, vol.~07,
  no.~03, pp. 183--194, July 2019.

\bibitem{Geiger_Lenz_Stiller_Urtasun_2013}
A.~Geiger, P.~Lenz, C.~Stiller, and R.~Urtasun, ``Vision meets robotics: The
  kitti dataset,'' \emph{The International Journal of Robotics Research},
  vol.~32, no.~11, p. 1231–1237, Sep 2013.

\bibitem{vis_compass}
C.~Giovannangeli and P.~Gaussier, ``Orientation system in robots: Merging
  allothetic and idiothetic estimations,'' \emph{13th International Conference
  on Advanced Robotics (ICAR07)}, 01 2007.

\bibitem{kohonen2012self}
T.~Kohonen, \emph{Self-organizing maps}.\hskip 1em plus 0.5em minus 0.4em\relax
  Springer Science \& Business Media, 2012, vol.~30.

\bibitem{Kuriscak_Marsalek_Stroffek_Toth_2015}
E.~Kuriscak, P.~Marsalek, J.~Stroffek, and P.~G. Toth, ``Biological context of
  hebb learning in artificial neural networks, a review,''
  \emph{Neurocomputing}, vol. 152, p. 27–35, Mar 2015.

\bibitem{lowryVisualPlaceRecognition2016}
S.~Lowry, N.~Sunderhauf, P.~Newman, J.~J. Leonard, D.~Cox, P.~Corke, and M.~J.
  Milford, ``\BIBforeignlanguage{en}{Visual {Place} {Recognition}: {A}
  {Survey}},'' \emph{\BIBforeignlanguage{en}{ieee transactions on robotics}},
  vol.~32, no.~1, p.~19, 2016.

\bibitem{RobotCarDatasetIJRR}
\BIBentryALTinterwordspacing
W.~Maddern, G.~Pascoe, C.~Linegar, and P.~Newman, ``{1 Year, 1000km: The Oxford
  RobotCar Dataset},'' \emph{The International Journal of Robotics Research
  (IJRR)}, vol.~36, no.~1, pp. 3--15, 2017. [Online]. Available:
  \url{http://dx.doi.org/10.1177/0278364916679498}
\BIBentrySTDinterwordspacing

\bibitem{Mallat_Zhifeng_Zhang_1993}
S.~Mallat and Z.~Zhang, ``Matching pursuits with time-frequency dictionaries,''
  \emph{IEEE Transactions on Signal Processing}, vol.~41, no.~12, p.
  3397–3415, Dec 1993.

\bibitem{Olshausen_Field_2004}
B.~Olshausen and D.~Field, ``Sparse coding of sensory inputs,'' \emph{Current
  Opinion in Neurobiology}, vol.~14, no.~4, p. 481–487, Aug 2004.

\bibitem{Paik_Ringach_2012}
S.-B. Paik and D.~L. Ringach, ``Link between orientation and retinotopic maps
  in primary visual cortex,'' \emph{Proceedings of the National Academy of
  Sciences}, vol. 109, no.~18, p. 7091–7096, May 2012.

\bibitem{Perrinet_2010}
L.~U. Perrinet, ``Role of homeostasis in learning sparse representations,''
  \emph{Neural Computation}, vol.~22, no.~7, p. 1812–1836, Jul 2010.

\bibitem{Sermanet_Eigen_Zhang_Mathieu_Fergus_LeCun_2014}
\BIBentryALTinterwordspacing
P.~Sermanet, D.~Eigen, X.~Zhang, M.~Mathieu, R.~Fergus, and Y.~LeCun,
  ``Overfeat: Integrated recognition, localization and detection using
  convolutional networks,'' \emph{arXiv:1312.6229 [cs]}, Feb 2014, arXiv:
  1312.6229. [Online]. Available: \url{http://arxiv.org/abs/1312.6229}
\BIBentrySTDinterwordspacing

\bibitem{Serre_Wolf_Bileschi_Riesenhuber_Poggio_2007}
T.~Serre, L.~Wolf, S.~Bileschi, M.~Riesenhuber, and T.~Poggio, ``Robust object
  recognition with cortex-like mechanisms,'' \emph{IEEE Transactions on Pattern
  Analysis and Machine Intelligence}, vol.~29, no.~3, p. 411–426, Mar 2007.

\bibitem{Spanne_Jorntell_2015}
A.~Spanne and H.~Jörntell, ``Questioning the role of sparse coding in the
  brain,'' \emph{Trends in Neurosciences}, vol.~38, no.~7, p. 417–427, Jul
  2015.

\bibitem{jessicavanbrummelenAutonomousVehiclePerception2018}
J.~Van~Brummelen, M.~O’Brien, D.~Gruyer, and H.~Najjaran,
  ``\BIBforeignlanguage{en}{Autonomous vehicle perception: {The} technology of
  today and tomorrow},'' \emph{\BIBforeignlanguage{en}{Transportation Research
  Part C: Emerging Technologies}}, vol.~89, pp. 384--406, Apr. 2018.

\bibitem{Voulodimos_Doulamis_Doulamis_Protopapadakis}
A.~Voulodimos, N.~Doulamis, A.~Doulamis, and E.~Protopapadakis, ``Deep learning
  for computer vision: A brief review,'' \emph{Computational Intelligence and
  Neuroscience}, vol. 2018, pp. 1--13, 02 2018.

\bibitem{yurtseverSurveyAutonomousDriving2019}
\BIBentryALTinterwordspacing
E.~Yurtsever, J.~Lambert, A.~Carballo, and K.~Takeda, ``A survey of autonomous
  driving: Common practices and emerging technologies,'' \emph{CoRR}, vol.
  abs/1906.05113, 2019. [Online]. Available:
  \url{http://arxiv.org/abs/1906.05113}
\BIBentrySTDinterwordspacing

\bibitem{Zaffar_review_2020}
\BIBentryALTinterwordspacing
M.~Zaffar, S.~Ehsan, M.~Milford, D.~Flynn, and K.~McDonald-Maier, ``Vpr-bench:
  An open-source visual place recognition evaluation framework with
  quantifiable viewpoint and appearance change,'' \emph{arXiv:2005.08135 [cs]},
  May 2020, arXiv: 2005.08135. [Online]. Available:
  \url{http://arxiv.org/abs/2005.08135}
\BIBentrySTDinterwordspacing

\bibitem{Zaffar_CoHog_2020}
M.~Zaffar, S.~Ehsan, M.~Milford, and K.~McDonald-Maier, ``Cohog: A
  light-weight, compute-efficient, and training-free visual place recognition
  technique for changing environments,'' \emph{IEEE Robotics and Automation
  Letters}, vol.~5, no.~2, p. 1835–1842, Apr 2020.

\bibitem{zenoReviewNeurobiologicallyBased2016}
P.~J. Zeno, S.~Patel, and T.~M. Sobh, ``Review of neurobiologically based
  mobile robot navigation system research performed since 2000,'' \emph{Journal
  of Robotics}, vol. 2016, 2016.

\end{thebibliography}


\end{document}